\documentclass[wcp]{jmlr}

\usepackage{enumerate}
\usepackage{bbold}
\usepackage{booktabs}
\usepackage{natbib}
\usepackage[utf8]{inputenc}
\usepackage[super]{nth}
\jmlryear{2014}
\jmlrworkshop{Neural Connectomics Workshop}

\title{Simple connectome inference from partial correlation statistics in calcium imaging}

  \author{\Name{Antonio Sutera},
   \Name{Arnaud Joly},
   \Name{Vincent François-Lavet}, \Email{a.sutera@ulg.ac.be}\\
   \Name{Zixiao Aaron Qiu},
   \Name{Gilles Louppe},
   \Name{Damien Ernst}\and\Name{Pierre Geurts}
    \\
   \addr Department of EE and CS \& GIGA-R, University of Li\`ege, Belgium}

\begin{document}

\maketitle

\vspace{-1cm}
\begin{abstract}
In this work, we propose a simple yet effective solution to the problem of
connectome inference in calcium imaging data. The proposed algorithm consists of
two steps. First, processing the raw signals to detect neural peak activities.
Second, inferring the degree of association between neurons from partial
correlation statistics. This paper summarises the methodology that led us to
win the Connectomics Challenge, proposes a simplified version of our method, and
finally compares our results with respect to other inference methods.
\end{abstract}


\section{Introduction}\label{sec:intro}

The human brain is a complex biological organ made of about 100 billion of
neurons, each connected to, on average, 7,000 other neurons
\citep{pakkenberg2003aging}. Unfortunately, direct observation of the
connectome, the wiring diagram of the brain, is not yet technically feasible.
Without being perfect, calcium imaging currently allows for real-time and
simultaneous observation of neuron activity from thousands of neurons,
producing individual time-series representing their fluorescence intensity.
From these data, the connectome inference problem amounts to retrieving the
synaptic connections between neurons on the basis of the fluorescence time-series. This problem is difficult to solve because of experimental issues,
including masking effects (i.e., some of the neurons are not observed or
confounded with others), the low sampling rate of the optical device with
respect to the neural activity speed, or the slow decay of fluorescence.

Formally, the connectome can be represented as a directed graph $G=(V,E)$,
where $V$ is a set of $p$ nodes representing neurons, and $E \subseteq
\left\{(i, j) \in V \times V\right\}$ is a set of edges representing direct
synaptic connections between neurons. Causal interactions are expressed by the
direction of edges: $(i, j) \in E$ indicates that the state of neuron $j$ might
be caused by the activity of neuron $i$. In those terms,  the connectome
inference problem is formally stated as follows:  \textit{Given the sampled
observations $\{ x^t_i \in \mathbb{R} | i \in V, t = 1, \dots, T \}$ of $p$
neurons for $T$ time intervals, the goal is to infer the set $E$ of connections
in $G$.}

In this paper, we present a simplified - and almost as good - version of the
winning method\footnote{Code available at \url{https://github.com/asutera/kaggle-connectomics}} of the Connectomics
Challenge\footnote{\url{http://connectomics.chalearn.org}}, as a simple and
theoretically grounded approach based on signal processing techniques and
partial correlation statistics. The paper is structured as follows:
Section~\ref{sec:filter} describes the signal processing methods applied on
fluorescent calcium time-series; Section \ref{sec:inference} then presents the
proposed approach and its theoretical properties; Section~\ref{sec:results}
provides an empirical analysis and comparison with other network inference
methods, while finally, in Section~\ref{sec:conclusion} we discuss our work and
provide further research directions. Additionally,
Appendix~\ref{app:optimized} further describes, in full detail, our actual
winning method which gives slightly better results than the method presented in
this paper, at the cost of parameter tuning. Appendix~\ref{app:supp} provides supplementary results on other datasets.

\section{Signal processing} \label{sec:filter}

\begin{figure}
\centering
\subfigure[Raw signal]{\includegraphics[width=0.3\textwidth]{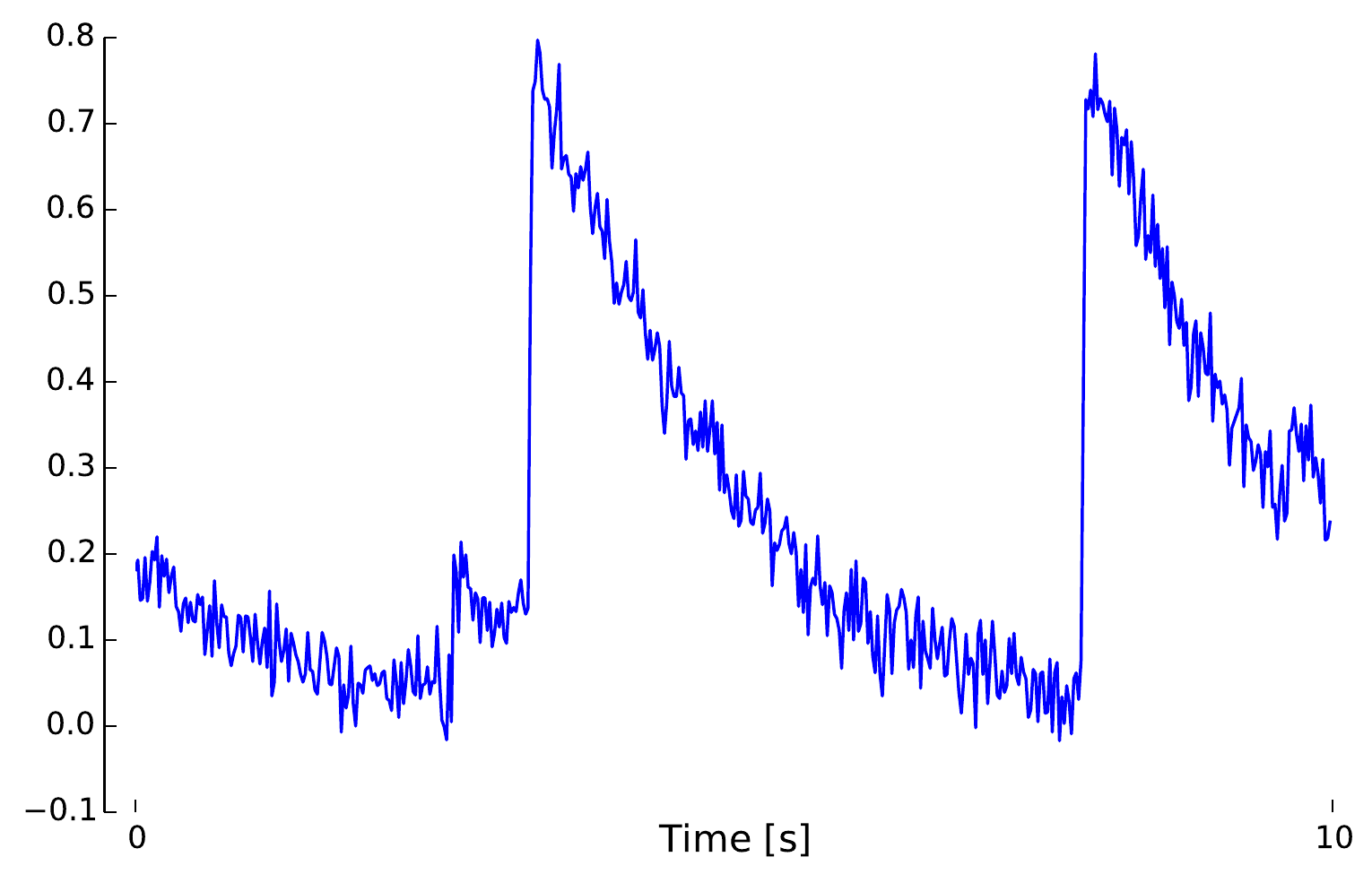} \label{fig:original_curve}}
\subfigure[Low-pass filter $f_1$]{\includegraphics[width=0.3\textwidth]{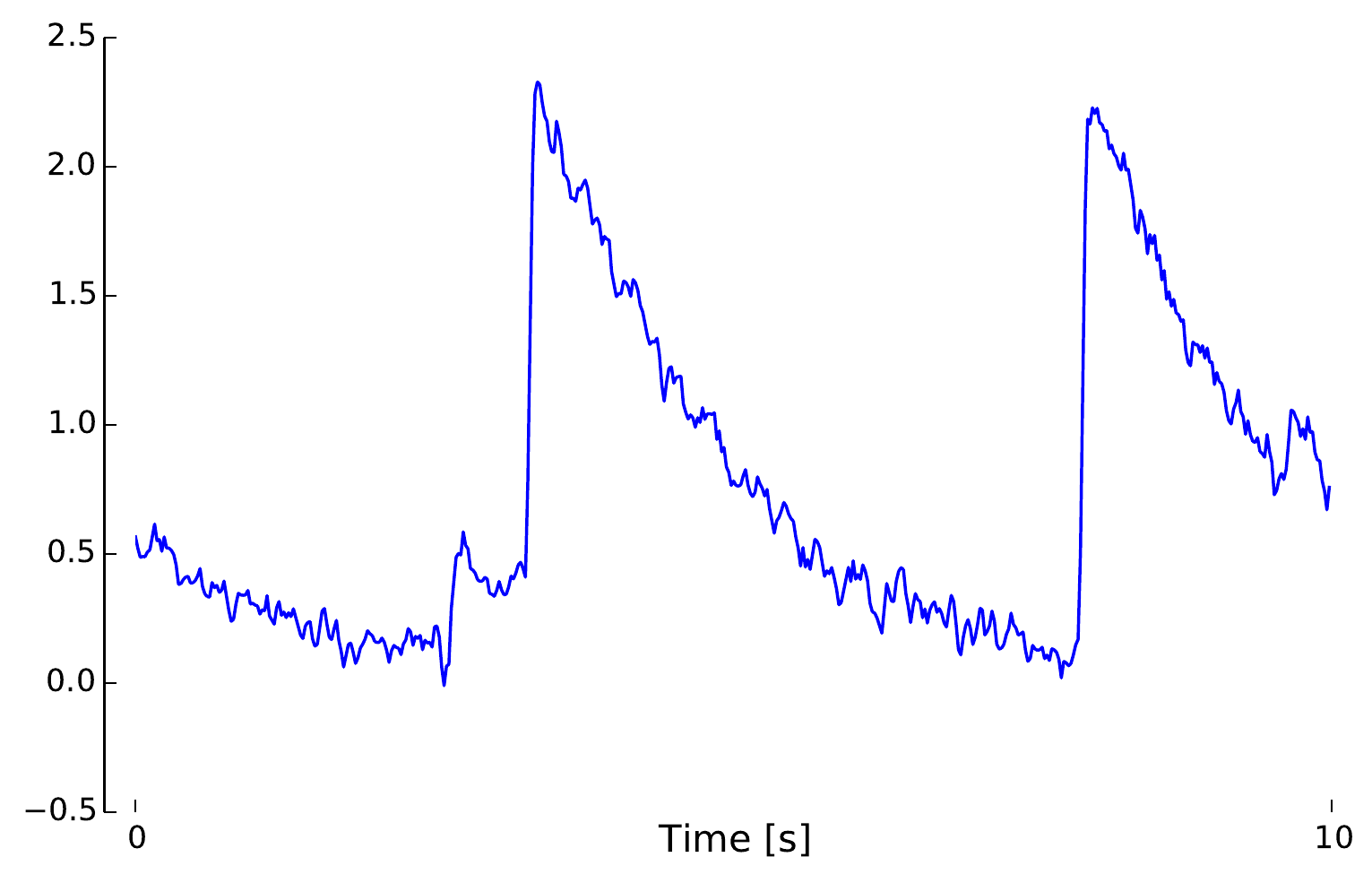} \label{fig:lp_curve}}
\subfigure[High-pass filter $g$]{\includegraphics[width=0.3\textwidth]{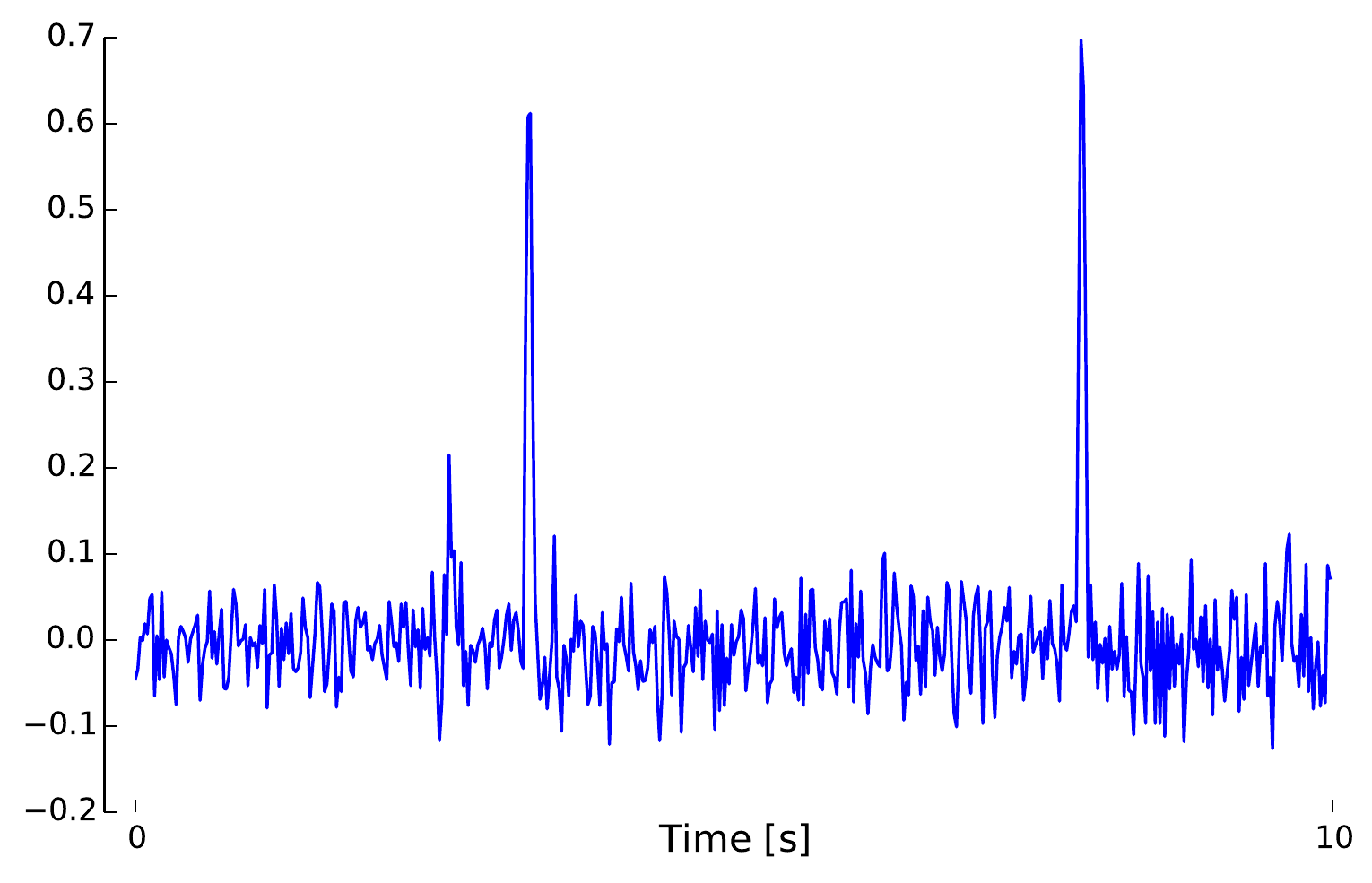} \label{fig:hp_curve}}\\
\subfigure[Hard-threshold filter $h$]{\includegraphics[width=0.3\textwidth]{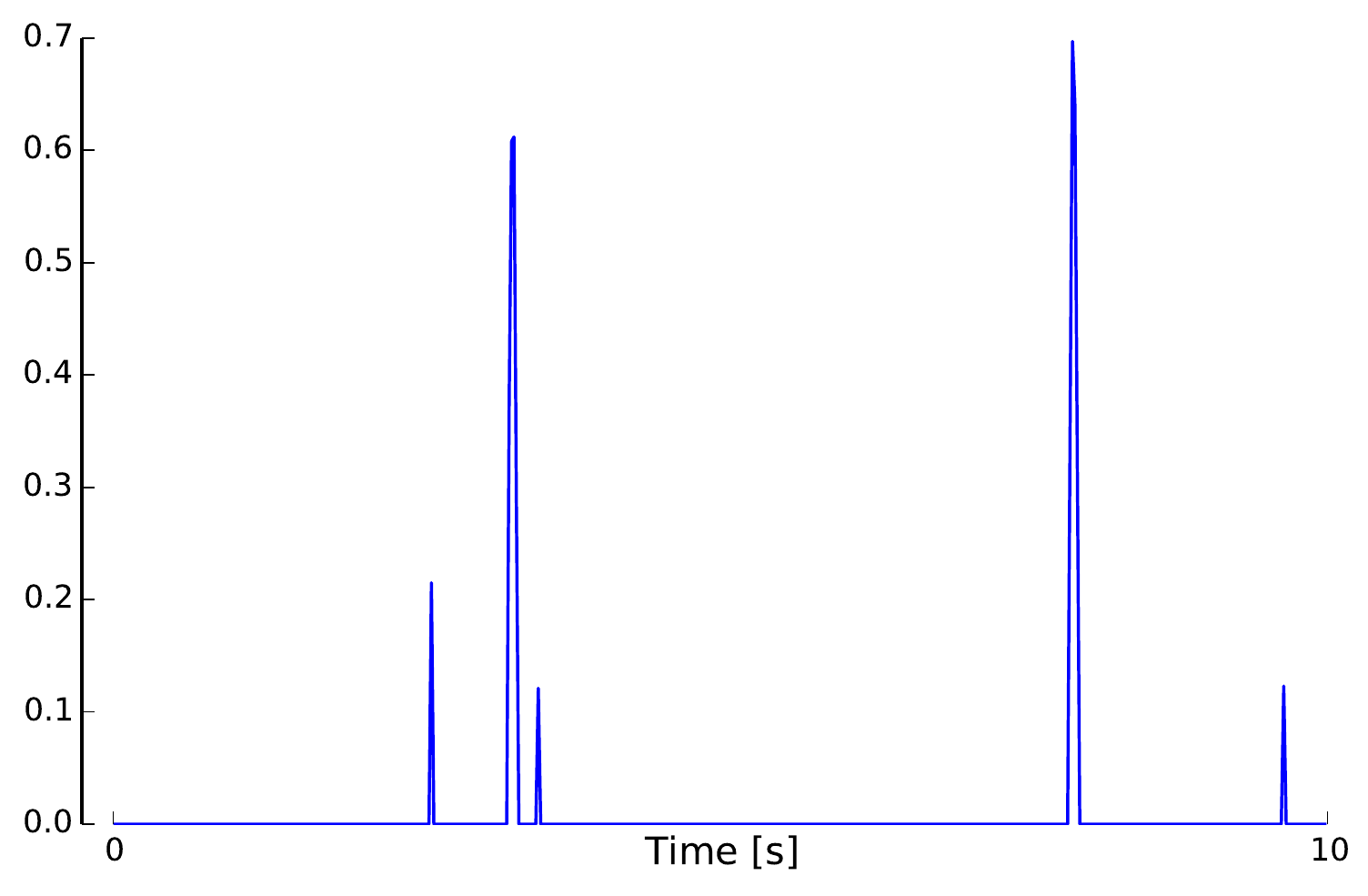} \label{fig:threshold_curve}}
\subfigure[Global regularization $w$]{\includegraphics[width=0.3\textwidth]{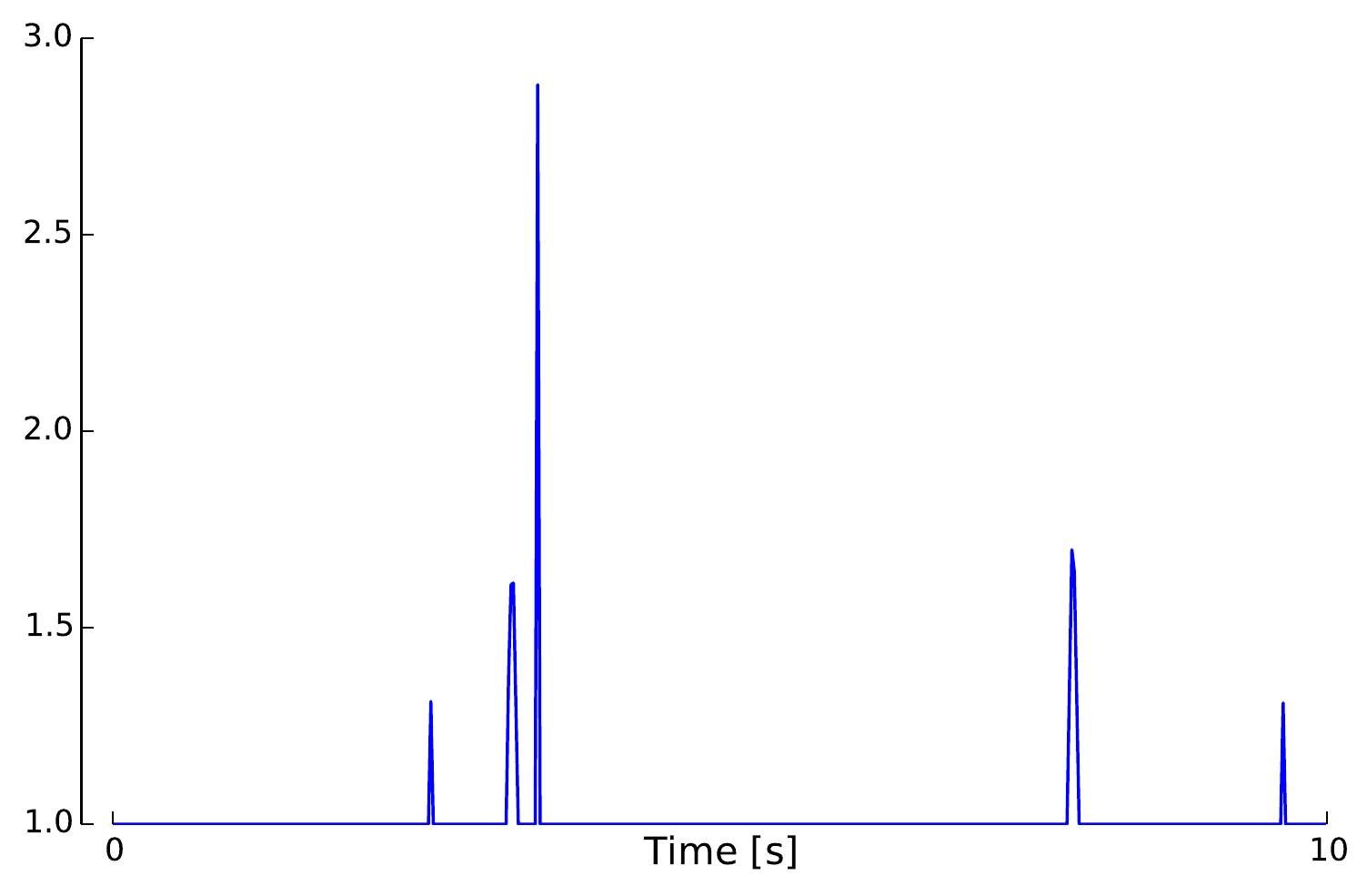} \label{fig:weight_curve}}
\caption{Signal processing pipeline for extracting peaks from the raw fluorescence data.}
\label{fig:filtered-signal}
\end{figure}

Under the simplifying assumption that neurons are on-off units, characterised
by short periods of intense activity, or peaks, and longer periods of
inactivity, the first part of our algorithm consists of cleaning the raw
fluorescence data.
More specifically, time-series are processed using standard
signal processing filters in order to : (i) remove noise mainly due to fluctuations independent of calcium, calcium fluctuations independent of spiking activity, calcium fluctuations in nearby tissues that have been mistakenly captured, or simply by the imaging process ; (ii) to account for fluorescence low decay ; and (iii) to reduce the importance of
high global activity in the network. The overall process is illustrated in
Figure~\ref{fig:filtered-signal}.

As Figure~\ref{fig:original_curve} shows, the raw
fluorescence signal is very noisy due to light scattering artifacts that
usually affect the quality of the recording~\citep{lichtman2011big}.
Accordingly, the first step of our pipeline is to smoothe the signal, using
one of the following low-pass filters for filtering out high frequency noise:
\begin{align}
f_1(x^t_i) &= x^{t-1}_i + x^t_i + x^{t+1}_i, \label{eq:symetric-median} \\
f_2(x^t_i) &= 0.4 x^{t-3}_i + 0.8 x^{t-2}_i + x^{t-1}_i + x_i^t.
\label{eq:weighted-asymetric-median}
\end{align}
These filters are standard in the signal processing field \citep{kaiser1977data, oppenheim1983signals}. For the purposes of illustration, the effect of the filter $f_1$ on the signal
is shown in Figure \ref{fig:lp_curve}.

Furthermore, short spikes, characterized by a high
frequency, can be seen as an indirect indicator of neuron communication, while low frequencies of the signal mainly correspond to the slow
decay of fluorescence. To have a signal that only has high magnitude around instances where the spikes occur, the second step of our pipeline transforms the time-series into its backward
difference
\begin{align}
g(x^{t}_{i}) &= x^{t}_i - x^{t-1}_i, \label{eq:high-pass-filter}
\end{align}
as shown in Figure \ref{fig:hp_curve}.

To filter out small variations in the signal obtained after applying the
function $g$, as well as to eliminate negative values, we use the following
hard-threshold filter
\begin{align}\label{eqn:hfilter}
h(x^{t}_i) &= x^{t}_i \mathbb{1}(x^{t}_i \geq \tau) \text{ with } \tau > 0,
\end{align}
yielding Figure \ref{fig:threshold_curve} where $\tau$ is the threshold parameter and $\mathbb{1}$ is the indicator function.
As can be seen, the processed signal only contains clean spikes.

The objective of the last step of our filtering procedure is to decrease the
importance of spikes that occur when there is high global activity in the
network with respect to  spikes that occur during normal activity. Indeed, we
have conjectured that when a large part of the network is firing, the rate at
which observations are made is not high enough to be able to detect
interactions, and that it would therefore be preferable to lower their
importance by changing their magnitude appropriately. Additionally, it is
well-known that neurons may also spike because of a high global activity
\citep{stetter2012model}. In such  context, detecting pairwise neuron
interactions from the firing activity is meaningless. As such,
the signal output by $h$ is finally applied to the following function
\begin{align}
 w(x^{t}_i) &= (x^{t}_i + 1 )^{1 + \frac{1}{\sum_{j} x^{t}_j}}, \label{eq:magnify-filter}
\end{align}
whose effect is to magnify the importance of spikes that occur in cases of low
global activity (measured by $\sum_{j} x^{t}_j$), as observed, for instance,
around $t=4\text{s}$ in Figure~\ref{fig:weight_curve}. Note the particular case where there
is no activity, i.e., $\sum_{j} x^{t}_j = 0$, is solved by setting $w(x^{t}_i)
= 1$.

To summarise, the full signal processing pipeline of our simplified approach is defined by the composed function $w \circ h \circ g \circ
f_1$ (resp. $f_2$). When applied to the raw signal of Figure
\ref{fig:original_curve}, it outputs the signal shown in Figure
\ref{fig:weight_curve}.

\section{Connectome inference from partial correlation statistics}
\label{sec:inference}

Our procedure to infer connections between neurons first assumes that
the (filtered) fluorescence concentrations of all $p$ neurons at each
time point can be modelled as a set of random variables $X = \{X_1,
\dots, X_p\}$ that are independently drawn from the same time-invariant
joint probability distribution $P_X$. 
As a consequence, our inference method does not exploit the time-ordering of the observations (although time-ordering is exploited by
the filters).

Given this assumption, we then propose to use as a measure of the
strength of the connection between two neurons $i$ and $j$, the
\textit{Partial correlation} coefficient $p_{i,j}$ between their corresponding
random variables $X_i$ and $X_j$, defined by:
\begin{equation}
p_{i,j} =
-\frac{\Sigma^{-1}_{ij}}{\sqrt{\Sigma^{-1}_{ii} \Sigma^{-1}_{jj}}}, \label{eq:inverse}
\end{equation}
where $\Sigma^{-1}$, known as the precision or concentration matrix, is the inverse of the covariance matrix $\Sigma$ of $X$. 
Assuming that the distribution $P_X$ is a multivariate Gaussian
distribution ${\cal N}(\mu,\Sigma)$, it can be shown that $p_{i,j}$ is
zero if and only if $X_i$ and $X_j$ are independent given all other
variables in $X$, i.e., $X_i \perp X_j|X^{-i,j}$ where $X^{-i,j}= X
\setminus\{X_i,X_j\}$. Partial correlation thus measures conditional
dependencies between variables ; therefore it should naturally only detect direct associations
between neurons and filter out spurious indirect effects. The interest
of partial correlation as an association measure has already been
shown for the inference of gene regulatory networks
\citep{de2004discovery,Schafer:2005}.
Note that the partial correlation statistic is symmetric
(i.e. $p_{i,j}=p_{j,i}$). Therefore, our approach cannot identify the
direction of the interactions between neurons. We will see in
Section~\ref{sec:results} why this only slightly affects its
performance, with respect to the metric used in the Connectomics
Challenge. 

Practically speaking, the computation of all $p_{i,j}$ coefficients using Equation
\ref{eq:inverse} requires the estimation of the covariance matrix $\Sigma$
and then computing its inverse. Given that typically we have more
samples than neurons, the covariance matrix can be inverted in a
straightforward way. We nevertheless obtained some improvement by
replacing the exact inverse with an approximation using only the $M$
first principal components \citep{bishop2006pattern} (with
$M=0.8 p$ in our experiments, see Appendix~\ref{app:pca}). 

Finally, it should be noted that the performance of our simple method appears to
be quite sensitive to the values of parameters (e.g., choice of $f_1$ or $f_2$
or the value of the threshold $\tau$) in the combined function of the
filtering and inferring processes. One approach, further referred
to as \textit{Averaged Partial correlation} statistics, for improving
its robustness is to average correlation statistics over various
values of the parameters, thereby reducing the variance of its
predictions. Further details about parameter selection are provided in
Appendix~\ref{app:optimized}.

\section{Experiments} \label{sec:results}
\paragraph{Data and evaluation metrics.}

We report here experiments on the \textit{normal-1,2,3}, and \textit{4}
datasets provided by the organisers of the Connectomics Challenge (see
Appendix~\ref{app:supp} for experiments on other datasets). Each of
these datasets is obtained from the simulation \citep{stetter2012model} of
different neural networks of 1,000 neurons and approximately 15,000 edges (i.e., a
network density of about 1.5\%). Each neuron is described by a calcium
fluorescence time-series of length $T=179500$. All inference methods compared
here provide a ranking of all pairs of neurons according to some association score. To assess the quality of this ranking, we compute both ROC and
precision-recall curves against the ground-truth network, which are represented by
the area under the curves and respectively  denoted AUROC and AUPRC. Only
the AUROC score was used to rank the challenge participants, but the precision-recall curve has been shown to be a more sensible metric for network
inference, especially when network density is small (see e.g.,
\cite{schrynemackers2013protocols}). Since neurons are not self-connected in
the ground-truth networks (i.e., $(i, i) \not \in E, \forall i \in V$), we
have manually set the score of such edges to the minimum possible association
score before computing ROC and PR curves.

\paragraph{Evaluation of the method.}

The top of Table \ref{tab:comparison} reports AUROC and AUPRC for all
four networks using, in each case, partial correlation with different
filtering functions. Except for the last two rows that use PCA, the
exact inverse of the covariance matrix was used in each case. These
results clearly show the importance of the filters. AUROC increases in
average from 0.77 to 0.93. PCA does not really affect AUROC scores, but
it significantly improves AUPRC scores. Taking the average over
various parameter settings gives an improvement of 10\% in AUPRC but
only a minor change in AUROC. The last row (``Full method'') shows the
final performance of the method specifically tuned for the challenge
(see Appendix \ref{app:optimized} for all details). Although this
tuning was decisive to obtain the best performance in the challenge,
it does not significantly improve either AUROC or AUPRC.

\begin{table}[t]

\caption{Top: Performance on \textit{normal-1,2,3,4} with partial correlation and different filtering functions.
Bottom: Performance on \textit{normal-1,2,3,4} with different methods.}
\label{tab:comparison}
\centering
\small
\begin{tabular}{| l | c c c c | c c c c |}
\hline
& \multicolumn{4}{c|}{AUROC} & \multicolumn{4}{c|}{AUPRC} \\
\textit{Method} $\backslash$ \textit{normal-} & \textit{1} & \textit{2} & \textit{3} & \textit{4} & \textit{1} & \textit{2} & \textit{3} & \textit{4} \\
\hline
\hline
No  filtering       					& 0.777 & 0.767 & 0.772 & 0.774 & 0.070 & 0.064 & 0.068 & 0.072\\
$ h \circ g \circ f_1$                  & 0.923 & 0.925 & 0.923 & 0.922 & 0.311 & 0.315 & 0.313 & 0.304\\
$ w \circ h \circ g \circ f_1$          & 0.931 & 0.929 & 0.928 & 0.926 & 0.326 & 0.323 & 0.319 & 0.303\\
+ PCA         							& 0.932 & 0.930 & 0.928 & 0.926 & 0.355 & 0.353 & 0.350 & 0.333\\
Averaging           					& 0.937 & 0.935 & 0.935 & 0.931 & 0.391 &  0.390 &  0.385 & 0.375\\
Full method           					& \textbf{0.943} & \textbf{0.942} & \textbf{0.942} & \textbf{0.939} & \textbf{0.403} & \textbf{0.404} & \textbf{0.398} & \textbf{0.388}\\
\hline
PC & 0.886 & 0.884 & 0.891 &  0.877 & 0.153 & 0.145 & 0.170 & 0.132\\
GTE & 0.890 & 0.893 & 0.894 & 0.873 & 0.171 & 0.174 & 0.197 & 0.142\\
GENIE3 & 0.892 & 0.891 & 0.887 & 0.887 & 0.232 & 0.221 & 0.237 & 0.215 \\
\hline
\end{tabular}
\end{table}

\paragraph{Comparison with other methods.}

\begin{figure}[t]
\centering
\subfigure[ROC curves]{\includegraphics[width=0.45\textwidth]{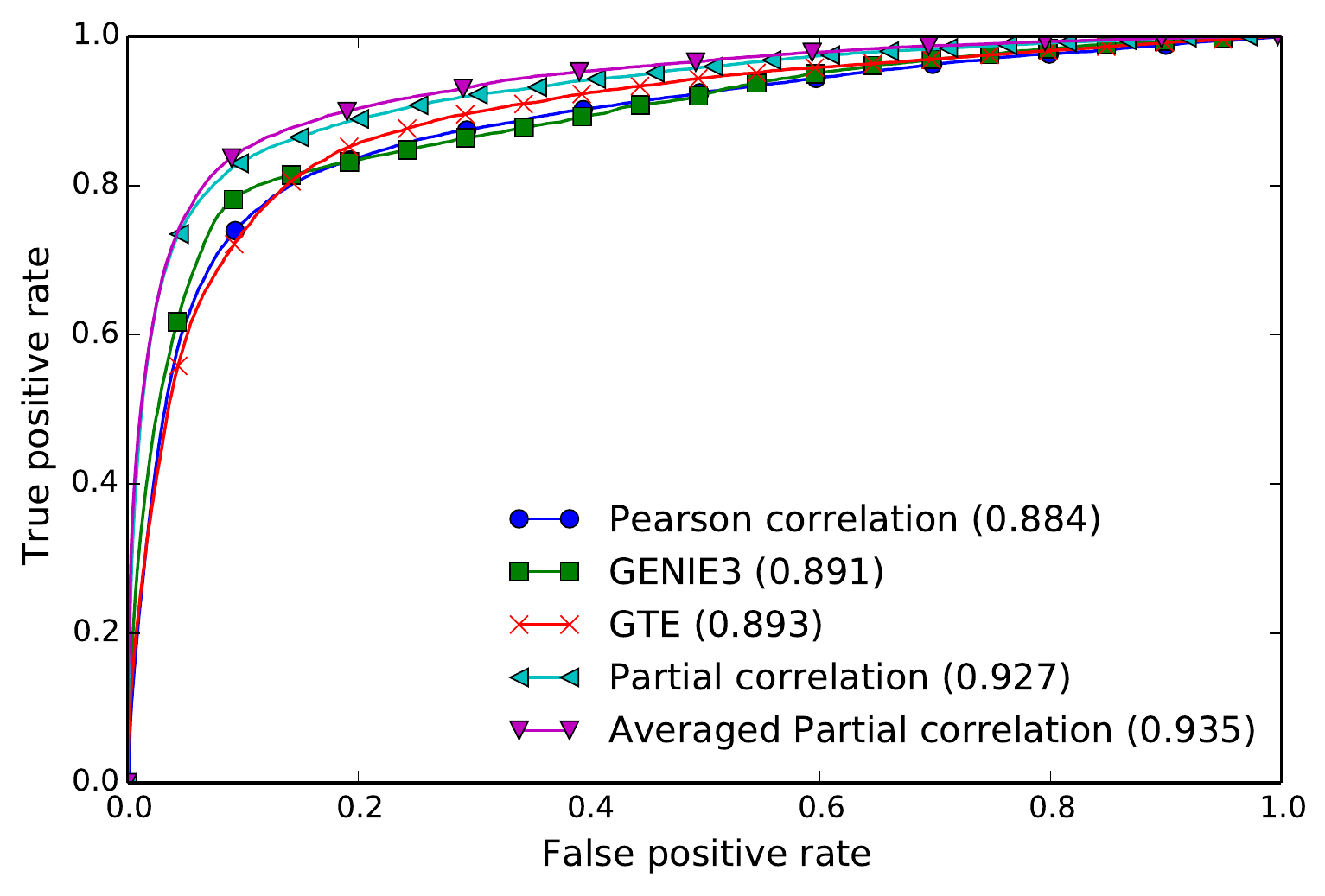} \label{fig:roc_curve}}
\subfigure[Precision-recall curves]{\includegraphics[width=0.45\textwidth]{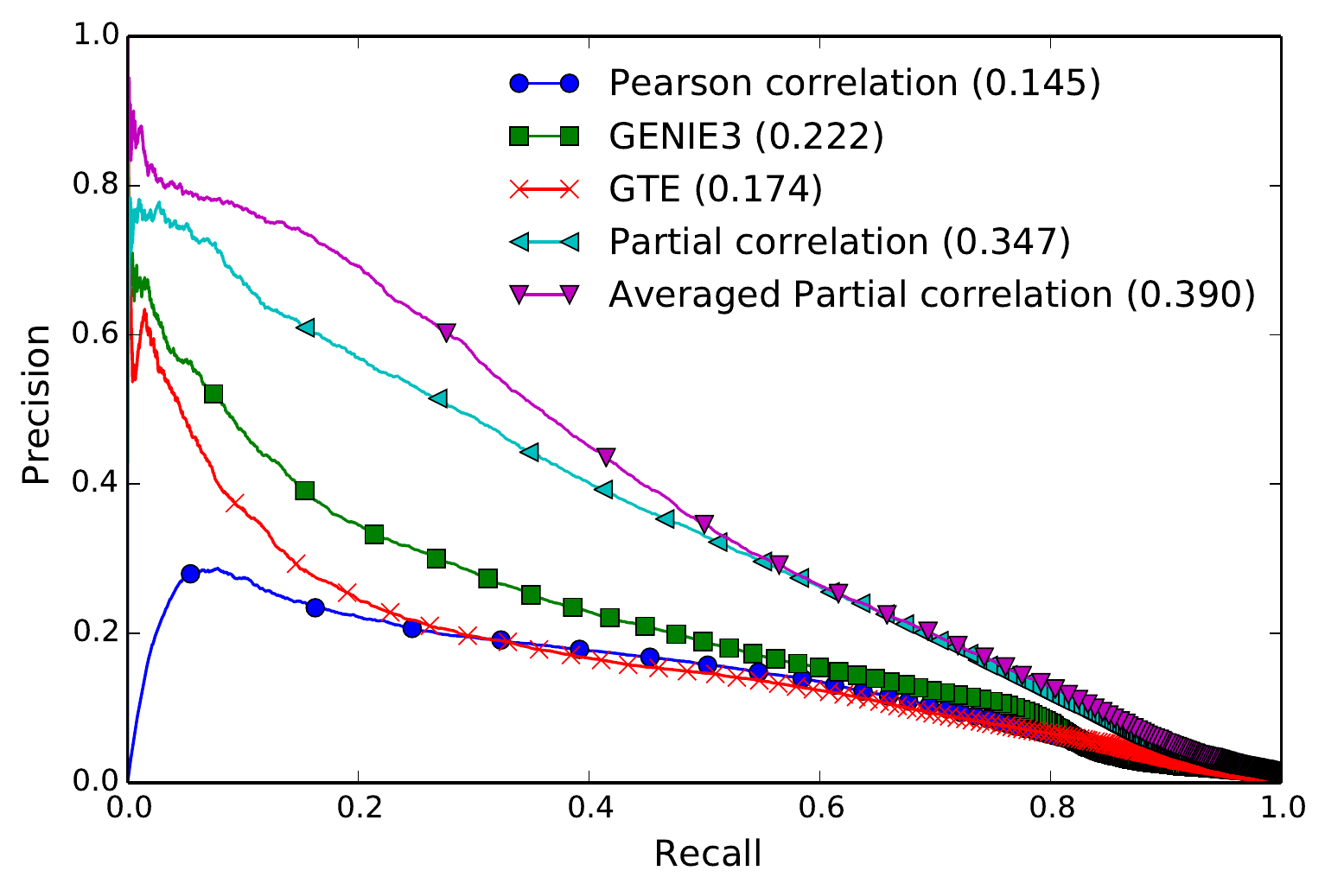} \label{fig:pr_curve}}
\caption{ROC (left) and PR (right) curves on \textit{normal-2} for the compared methods. Areas under the curves are reported in the legend.}
\label{fig:curves}
\end{figure}

At the bottom of Table \ref{tab:comparison}, we provide as a comparison the
performance of three other methods: standard (Pearson) correlation (PC),
generalised transfer entropy (GTE), and GENIE3. ROC and PR curves on the
\textit{normal-2} network are shown for all methods in Figure\ref{fig:curves}. Pearson correlation measures the unconditional linear
(in)dependence between variables and it should thus not be able to filter out
indirect interactions between neurons. GTE \citep{stetter2012model} was
proposed as a baseline for the challenge. This method builds on Transfer
Entropy to measure the association between two neurons. Unlike our approach, it
can predict the direction of the edges. GENIE3 \citep{huynhthu2010inferring} is
a gene regulatory network inference method that was the best performer in the
DREAM5 challenge \citep{marbach2012}. When transposed to neural networks, this
method uses the importance score of variable $X_i$ in a Random Forest model trying to
predict $X_j$ from all variables in $X\setminus X_j$ as a confidence score for the edge going from neuron $i$ to neuron
$j$. However, to reduce the
computational cost of this method, we had to limit each tree in the
Random Forest model to a maximum depth of 3. This constraint has a potentially
severe effect on the performance of this method with respect to the use of
fully-grown trees. PC and GENIE3 were applied to the time-series filtered using the functions $w\circ h\circ g$ and $h\circ g\circ f_1$ (which
gave the best performance), respectively. For GENIE3, we built 10,000 trees per neuron and we
used default settings for all other parameters (except for the maximal tree
depth). For GTE, we reproduced the exact same setting (conditioning level and
pre-processing) that was used by the organisers of the challenge.

Partial correlation and averaged partial correlation clearly outperform all
other methods on all datasets (see Table \ref{tab:comparison} and Appendix \ref{app:supp}). The
improvement is more important in terms of AUPRC than in terms of AUROC. As
expected, Pearson correlation performs very poorly in terms of AUPRC. GTE and
GENIE3 work much better, but these two methods are nevertheless clearly below
partial correlation. Among these two methods, GTE is slightly better in terms
of AUROC, while GENIE3 is significantly better in terms of AUPRC. Given that we
had to limit this latter method for computational reasons, these results are
very promising and a comparison with the full GENIE3 approach is certainly part
of our future works.

The fact that our method is unable to predict edge directions does not seem to
be a disadvantage with respect to GTE and GENIE3. Although partial correlation
scores each edge, and its opposite, similarly, it can reach precision values
higher than 0.5 (see Figure \ref{fig:curves}(b)), suggesting that it mainly ranks high
pairs of neurons that interact in both directions.  It is interesting also to
note that, on \textit{normal-2}, a method that perfectly predicts the
undirected network (i.e., that gives a score of $1$ to each pair $(i,j)$ such that
$(i,j)\in E$ or $(j,i)\in E$, and $0$ otherwise) already reaches an AUROC as high
as $0.995$ and an AUPRC of $0.789$.

\section{Conclusions} \label{sec:conclusion}

In this paper, we outlined a simple but efficient methodology for the problem
of connectome inference from calcium imaging data. Our approach consists of two
steps: (i) processing fluorescence data to detect neural peak activities; and
(ii) inferring the degree of association between neurons from partial
correlation statistics. Its simplified variant outperforms other
network inference methods while its optimized version proved to be the best method
on the Connectomics Challenge. Given its simplicity and good performance, we
therefore believe that the methodology presented in this work
would constitute a solid and easily-reproducible baseline for further work in
the field of connectome inference.

\paragraph{Acknowledgments}
A. Joly and G. Louppe are research fellows of the FNRS, Belgium.  A. Sutera is a
recipient of an FRIA fellowship of FRS-FNRS, Belgium. This work is supported by
PASCAL2 and the IUAP DYSCO, initiated by the Belgian State, Science Policy
Office.

\newpage
\clearpage

\bibliography{references}

\newpage
\clearpage

\appendix

\section{Description  of the ``Full method''}
\label{app:optimized}

This section provides a detailed description of the method specifically tuned
for the Connectomics Challenge. We restrict our description to the
differences with respect to the simplified method presented in the main
paper. Most parameters were tuned so as to maximize AUROC on the
\textit{normal-1} dataset and our design choices were validated by monitoring
the AUROC obtained by the 145 entries we submitted during the
challenge. Although the tuned method performs better than the simplified one on
the challenge dataset, we believe that the tuned method clearly overfits the
simulator used to generate the challenge data and that the simplified method
should work equally well on new independent datasets. We nevertheless provide
the tuned method here for reference purposes. Our implementation of the tuned
method is available at \url{https://github.com/asutera/kaggle-connectomics}.

This appendix is structured as follows: Section~\ref{sapp:signal} describes
the differences in terms of signal processing. Section~\ref{sapp:averaging}
then provides a detailed presentation of the averaging approach.
Section~\ref{sapp:connectome} presents an approach to correct the $p_{i,j}$
values so as to take into account the edge directionality. Finally,
Section~\ref{sapp:results} presents some experimental results to validate the
different steps of our proposal.

\subsection{Signal processing}
\label{sapp:signal}

In Section~\ref{sec:filter}, we introduced four filtering functions ($f$, $g$,
$h$, and $w$) that are composed in sequence (i.e., $w \circ h \circ g \circ
f$) to provide the signals from which to compute partial correlation
statistics. Filtering is modified as follows in the tuned method:

\begin{itemize}
\item In addition to $f_1$ and $f_2$ (Equations \ref{eq:symetric-median} and
  \ref{eq:weighted-asymetric-median}), two alternative low-pass filters $f_3$
  and $f_4$ are considered:
\begin{align}
f_3(x^t_i) &= x^{t-1}_i + x^{t}_i + x^{t+1}_i + x^{t+2}_i, \label{eq:asymetric-median-forward} \\
f_4(x^t_i) &=  x_i^t + x^{t+1}_i  + x^{t+2}_i + x^{t+3}_i. \label{eq:asymetric-median}
\end{align}
\item An additional filter $r$ is applied to smoothe differences in peak magnitudes
  that might remain after the application of the hard-threshold filter $h$:
\begin{align}
r(x^t_i) = (x_i^t)^c,
\end{align}
with $c=0.9$.
\item Filter $w$ is replaced by a more complex filter $w^*$ defined as:
\begin{align}
 w^*(x^{t}_i) &= {(x^{t}_i + 1 )^{\left (1 + \frac{1}{\sum_{j} x^{t}_j}\right )}}^{k(\sum_{j} x^{t}_j)}
\end{align}
where the function $k$ is a piecewise linear function optimised separately for
each filter $f_1$, $f_2$, $f_3$ and $f_4$ (see the implementation for full
details). Filter $w$ in the simplified method is a special case of $w^*$ with
$k(\sum_j x_j^t)=1$.
\end{itemize}
The pre-processed time-series are then obtained by the application of the
following function: $w^*\circ r \circ h \circ g \circ f_i$ (with $i=1$, 2, 3, or 4).

\subsection{Weighted average of partial correlation statistics}
\label{sapp:averaging}

As discussed in Section \ref{sec:inference}, the performance of the method (in
terms of AUROC) is sensitive to the value of the parameter $\tau$ of the
hard-threshold filter $h$ (see Equation \ref{eqn:hfilter}), and to the choice
of the low-pass filter (among $\{f_1, f_2, f_3, f_4\}$).
As in the simplified method, we have averaged the partial correlation statistics obtained for all the pairs $(\tau,\mbox{low-pass filter}) \in \{0.100,0.101,\ldots,0.210\}\times \{f_1, f_2, f_3, f_4\}$.

Filters $f_1$ and $f_2$ display similar performances and thus were given similar
weights (i.e., resp. $0.383$ and $0.345$). These weights were chosen equal to the weights selected for the simplified method. In contrast, filters $f_3$
and $f_4$ turn out, individually, to be less competitive and were therefore given
less importance in the weighted average (i.e., resp. $0.004$ and $0.268$). Yet, as further shown in
Section~\ref{sapp:results}, combining all $4$ filters proves to marginally
improve performance with respect to using only $f_1$ and $f_2$.

\subsection{Prediction of edge orientation}
\label{sapp:connectome}

Partial  correlation  statistics is  a  symmetric  measure, while  the
connectome is a directed graph. It  could thus be beneficial to try to
predict edge orientation. In this section, we present an heuristic that
modifies the  $p_{ij}$ computed  by the  approach described  before which
takes into account directionality.

This approach is based on the following
observation. The rise of fluorescence of a neuron indicates its
activation. If another neuron is activated after a slight delay, this
could be a consequence of the activation of the first neuron and
therefore indicates a directed link in the connectome from the first to
the second neuron.  Given this observation, we have computed the following term for every
pair $(i,j)$:
\begin{align}
s_{i,j} = \sum_{t=1}^{T - 1} \mathbb{1}((x_j^{t+1} - x_i^t) \in \left[\phi_1, \phi_2\right])
\end{align}
that could be interpreted as an image of the  number of times
that neuron $i$ activates neuron $j$. $\phi_1$ and $\phi_2$ are
parameters whose values have been chosen in our experiments equal to
$0.2$ and $0.5$, respectively. Their role is to
define when the difference between $x_j^{t+1}$  and $x_i^t$ can
indeed be assimilated to an event for which neuron $i$ activates neuron
$j$.

Afterwards, we have computed the difference between $s_{i,j}$ and
$s_{j,i}$, that we call $z_{i,j}$, and used this difference to modify  $p_{i,j}$ and
$p_{j,i}$ so as to take into account directionality. Naturally, if
$z_{i,j}$ is greater  (smaller) than $0$, we may conclude that should there  be an
edge between $i$ and $j$, then this edge would have to be oriented
from $i$ to $j$ ($j$ to $i$).

This suggests the new association matrix $r$:
\begin{align}
r_{i,j} =  \mathbb{1}(z_{i,j} > \phi_3)  *  p_{i,j}
\end{align}
where $\phi_3 >0$ is another parameter. We discovered that this new
matrix $r$ was not providing good results, probably due to the fact that
directivity was not rewarded well enough in the challenge.

This has lead us to investigate other ways for exploiting the
information about directionality contained in the matrix $z$. One of
those ways that gave good performance was to use as an association
matrix:
\begin{align}
q_{i,j} = weight * p_{i,j} + (1-weight) * z_{i,j}
\label{eqn:qij}
\end{align}
with  $weight$ chosen close to 1 ($weight=0.997$). Note that with
values for $weight$ close to 1,   matrix $q$ only uses the
information to a minimum about directivity contained in $z$ to modify the  partial
correlation matrix $p$. We tried smaller values for $weight$ but those
provided poorer results.

It was  this association matrix $q_{i,j}$ that actually led to the
best results of the challenge, as shown in Table \ref{tab:directivity}
of Section~\ref{sapp:results}.

\subsection{Experiments}
\label{sapp:results}

\paragraph{On the interest of low-pass filters $f_3$ and $f_4$.}

As reported in Table~\ref{tab:f3f4}, averaging over all low-pass filters leads
to better AUROC scores than averaging over only two low-pass filters, i.e., $f_1$ and
$f_2$. However this slightly reduces AUPRC.

\begin{table}[ht]
\caption{Performance on \textit{normal-1, 2, 3, or 4} with partial correlation with different averaging approaches.}
\label{tab:f3f4}
\centering
\small
\begin{tabular}{| l | c c c c | c c c c |}
\hline
& \multicolumn{4}{c|}{AUROC} & \multicolumn{4}{c|}{AUPRC} \\
\textit{Averaging} $\backslash$ \textit{normal-} & \textit{1} & \textit{2} & \textit{3} & \textit{4} & \textit{1} & \textit{2} & \textit{3} & \textit{4} \\
\hline
\hline
 with $f_1$, $f_2$ & 0.937 & 0.935 & 0.935 & 0.931 & 0.391 &  \textbf{0.390} &  0.385 & \textbf{0.375}  \\
 with $f_1$, $f_2$, $f_3$, $f_4$ & \textbf{0.938} & \textbf{0.936} & \textbf{0.936} & \textbf{0.932} & 0.391 & 0.389 & 0.385 & 0.374\\
\hline
\end{tabular}
\end{table}

\paragraph{On the interest of using matrix $q$ rather than $p$ to take into account directivity.}

Table~\ref{tab:dir} compares AUROC and AUPRC with or without correcting the $p_{i,j}$ values according to Equation \ref{eqn:qij}. Both AUROC and AUPRC are (very slightly) improved by using information about directivity.

\begin{table}[ht]\label{tab:dir}
\caption{Performance on \textit{normal-1,2,3,4} of ``Full Method'' with and
  without using information about directivity.}
\label{tab:directivity}

\centering
\small
\begin{tabular}{| l | c c c c | c c c c |}
\hline
& \multicolumn{4}{c|}{AUROC} & \multicolumn{4}{c|}{AUPRC} \\
\textit{Full method} $\backslash$ \textit{normal-} & \textit{1} & \textit{2} & \textit{3} & \textit{4} & \textit{1} & \textit{2} & \textit{3} & \textit{4} \\
\hline
\hline
 Undirected & 0.943 & 0.942 & 0.942 & 0.939 & 0.403 & 0.404 & 0.398 & 0.388  \\
  Directed & \textbf{0.944} & \textbf{0.943} & 0.942 & \textbf{0.940} & \textbf{0.404} & \textbf{0.405} & \textbf{0.399} & \textbf{0.389}\\
\hline
\end{tabular}
\end{table}

\section{Supplementary results} \label{app:supp}

In this appendix we report the performance of the different methods compared
in the paper on 6 additional datasets provided by the Challenge
organisers. These datasets, corresponding each to networks of 1,000 neurons, are similar to
the \textit{normal} datasets except for one feature:
\begin{description}
\item[lowcon:] Similar network but on average with a lower number of connections per neuron.
\item[highcon:] Similar network but on average with a higher number of connections per neuron.
\item[lowcc:] Similar network but on average with a lower clustering coefficient.
\item[highcc:] Similar network but on average with a higher clustering coefficient.
\item[normal-3-highrate:] Same topology as \textit{normal-3} but with a higher firing frequency, i.e., with highly active neurons.
\item[normal-4-lownoise:] Same topology as \textit{normal-4} but with a better signal-to-noise ratio.
\end{description}

The results of several methods applied to these 6 datasets are provided in
Table~\ref{tab:results_appendix}. They confirm what we observed on the
\textit{normal} datasets. Average partial correlation and its tuned variant,
i.e.,``Full method'', clearly outperform other network inference methods on all
datasets. PC is close to GENIE3 and GTE, but still slightly worse. GENIE3
performs better than GTE most of the time. Note that the "Full method" reported in this table does not use Equation \ref{eqn:qij} to slightly correct the values of $p_{i,j}$  to take into account directivity.

\begin{table}[h]
\caption{Performance (top: AUROC, bottom: AUPRC) on specific datasets with different methods.}
\label{tab:results_appendix}
\centering
\small
\begin{tabular}{| l | c c c c c c |}
\hline
& \multicolumn{6}{c|}{AUROC}\\
\textit{Method} $\backslash$ \textit{normal-} & \textit{lowcon} & \textit{highcon} & \textit{lowcc} & \textit{highcc} & \textit{3-highrate} & \textit{4-lownoise} \\
\hline
\hline
Averaging     & 0.947 & 0.943 & 0.920 & 0.942 & 0.959 & 0.934 \\
Full method   & \textbf{0.955} & \textbf{0.944} &  \textbf{0.925} & \textbf{0.946} & \textbf{0.961} & \textbf{0.941} \\
PC & 0.782 & 0.920 &  0.846 & 0.897  & 0.898  & 0.873 \\
GTE & 0.846 & 0.905 & 0.848 & 0.899 & 0.905 & 0.879\\
GENIE3 & 0.781 &  0.924 & 0.879 & 0.902 & 0.886 &  0.890 \\ \hline \hline
& \multicolumn{6}{c|}{AUPRC}\\ \hline
Averaging     & 0.320 & 0.429 & 0.262 & 0.478 & 0.443 & 0.412 \\
Full method   & \textbf{0.334} & \textbf{0.413} &  \textbf{0.260} & \textbf{0.486} & \textbf{0.452} & \textbf{0.432}\\
PC & 0.074 & 0.218 & 0.082 & 0.165  & 0.193 & 0.135 \\
GTE & 0.094 & 0.211 & 0.081 & 0.165 & 0.210 & 0.144\\
GENIE3 & 0.128 & 0.273 & 0.116 & 0.309 & 0.256 & 0.224\\ \hline
\end{tabular}
\end{table}

\section{On the selection of the number of principal components}
\label{app:pca}

The (true) network, seen as a matrix, can be decomposed through a singular value decomposition (SVD) or principal component
analysis (PCA), so as to respectively determine a set of independent linear combinations of the
variable \citep{alter2000singular}, or a reduced set of linear
combinations combine, which then maximize the explained variance of the data
\citep{jolliffe2005principal}. Since SVD and PCA are related, they can be defined by the same goal: both aim at finding a reduced set of neurons, known as components, whose activity can explain the rest of the network.

\begin{figure}[t]
\centering
\subfigure[PCA]{\includegraphics[width=0.45\textwidth]{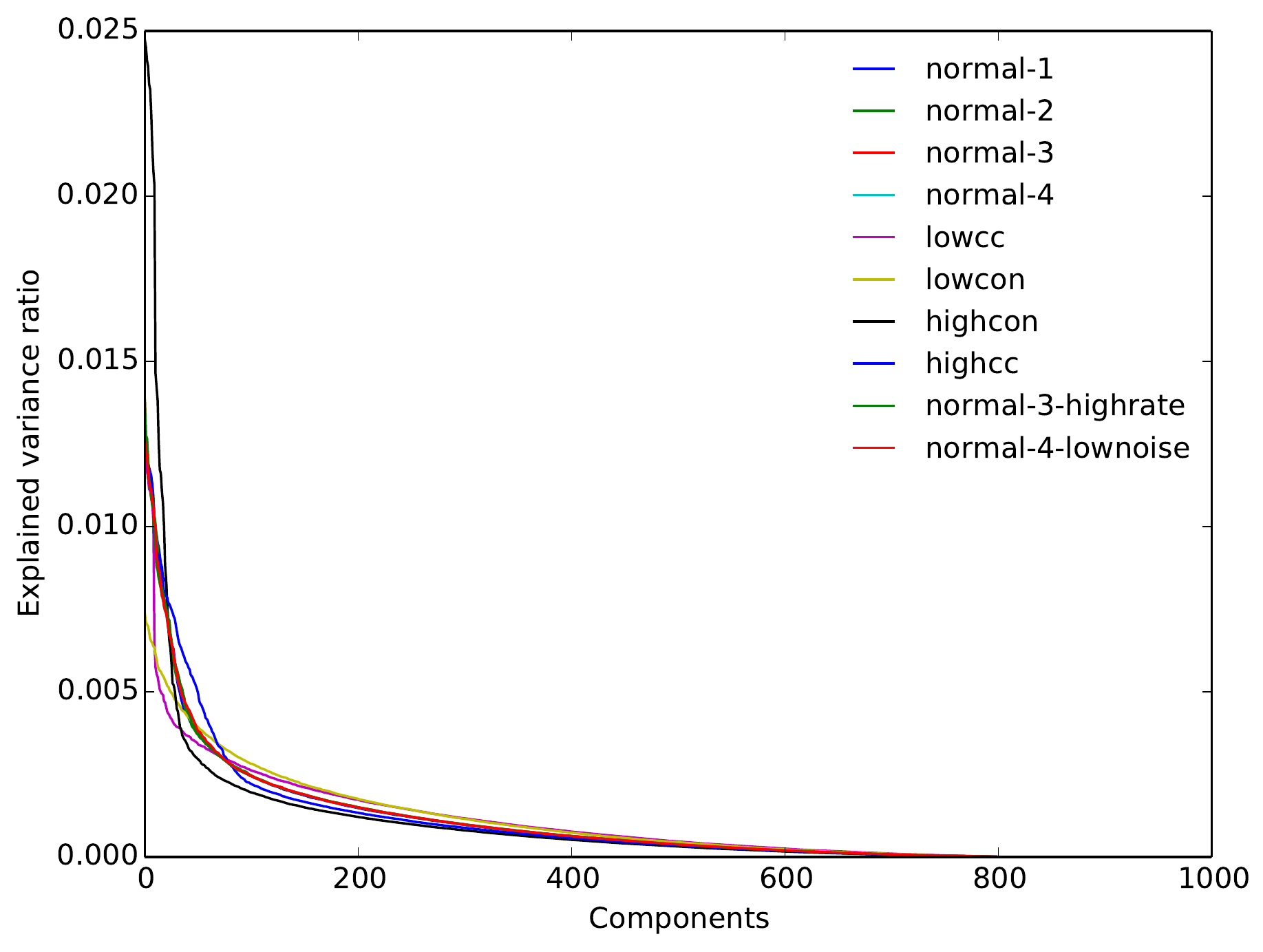} \label{fig:PCA}}
\subfigure[SVD]{\includegraphics[width=0.45\textwidth]{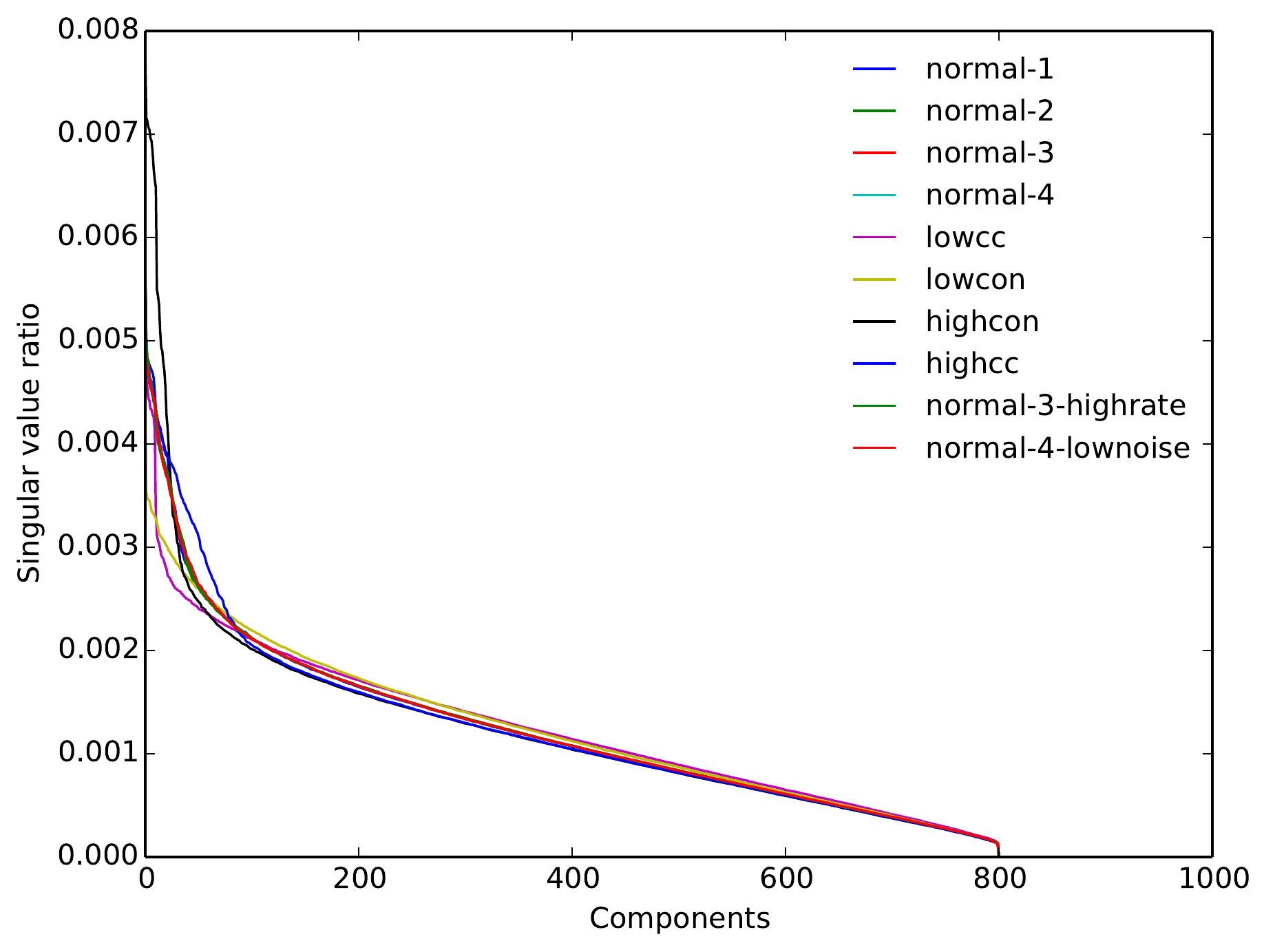} \label{fig:SVD}}
\caption{Explained variance ratio by number of principal components (left) and singular value ratio by number of principal components (right) for all networks.}
\label{fig:pca}
\end{figure}

The distribution of compoment eigen values obtained from PCA and SVD decompositions can be studied by sorting them in descending order of magnitude, as illustrated in Figure~\ref{fig:pca}. It can be seen that some component eigen values are zero, implying that the behaviour of the network could be explained by a subset of neurons because of the 
redundancy and relations between the neurons. For all datasets, the eigen value distribution is exactly the same. 

In the context of the challenge, we observe that only $800$ components seem to be necessary and we exploit this when computing partial correlation statistics. Therefore, the value of the parameter $M$ is immediate and should be clearly set to $800$ ($=0.8p$).

Note that if the true network is not available, similar decomposition analysis could be carried on the inferred network, or on the data directly.

\section{Summary Table}

\begin{table}[h]
 \caption{Connectomics Challenge summary.}
 \label{tab:summary}
 \centering
 \small
 \begin{tabular}{| l|c|} \hline
 Team Name & The AAAGV Team \\ \hline
 Private leaderboard position & \nth{1} \\ \hline
 Private leaderboard performance & 0.94161 \\ \hline
 Private leaderboard performance of the winner & idem \\ \hline
 \end{tabular}
 \end{table}

\end{document}